\documentclass{article} % For LaTeX2e
\usepackage{iclr2023_conference,times}
\usepackage{graphicx}
\usepackage[hidelinks]{hyperref}
\usepackage{url}

\title{Improving extreme weather events detection with light-weight
neural networks}
\author{Romain Lacombe\\
Stanford University\\
Plume Labs\\
\texttt{rlacombe@stanford.edu}
\And 
Hannah Grossman \hspace{3.64cm}\mbox{}\\ 
Stanford University\\
\texttt{hlg@stanford.edu  }
\AND Lucas Hendren\\
Stanford University\\
\texttt{hendren@stanford.edu}
\And David Lüdeke\\
Stanford University\\
\texttt{dludeke@stanford.edu}}

\iclrfinalcopy % Uncomment for camera-ready version, but NOT for submission.
\begin{document}

\maketitle

\begin{abstract}
% The abstract paragraph should be indented 1/2~inch (3~picas) on both left and
% right-hand margins. Use 10~point type, with a vertical spacing of 11~points.
% The word \textsc{Abstract} must be centered, in small caps, and in point size 12. Two
% line spaces precede the abstract. The abstract must be limited to one
% paragraph.
To advance automated detection of extreme weather events, which are increasing in frequency and intensity with climate change, we explore modifications to a novel light-weight Context Guided convolutional neural network architecture trained for semantic segmentation of tropical cyclones and atmospheric rivers in climate data. Our primary focus is on tropical cyclones, the most destructive weather events, for which current models show limited performance. We investigate feature engineering, data augmentation, learning rate modifications, alternative loss functions, and architectural changes. In contrast to previous approaches optimizing for intersection over union, we specifically seek to improve recall to penalize under-counting and prioritize identification of tropical cyclones. We report success through the use of weighted loss functions to counter class imbalance for these rare events. We conclude with directions for future research on extreme weather events detection, a crucial task for prediction, mitigation, and equitable adaptation to the impacts of climate change.

\end{abstract}

\section{Introduction}	

Climate action failure and extreme weather are two of the most severe global risks today \citep{IPCC, WEF}. Tropical cyclones, the most destructive extreme weather events \citep{NOAA}, have a rising and disproportionate impact on low and medium income countries (LMICs), yet research into their effects focuses mostly on high-income countries \citep{TC_LMICs}. Studies of extreme weather and climate change rely on heuristics or expert judgment to label data which leads to an inequitable global scientific focus, as well as discrepancies in predicted frequency, intensity, and attribution estimates. Improving automated detection of extreme weather events is thus paramount to fair attribution of climate loss and damages \citep{Attribution}, and to develop the early warning and detection systems that will be critical for equitable adaption to climate change \citep{IPCC, econ}.

Since 2020, deep learning has shown great promise for semantic segmentation of weather patterns in climate simulation data \citep{ClimateNet}. However, initial approaches have relied on complex architectures and hard to train models with very large numbers of parameters. A key area of research is the application of lighter-weight neural networks to semantic segmentation of tropical cyclones (TC) and atmospheric rivers (AR) \citep{NeurIPS2020}.

Here we explore the application of the light-weight Context Guided convolutional neural network (CGNet) architecture to semantic segmentation of tropical cyclones in climate data. Input to our model is hand-labeled climate simulation data with channels that contain key atmospheric variables such as wind speed, moisture content, and atmospheric pressure for different time steps, latitudes, and longitudes. The output is a segmentation mask where each pixel takes a value corresponding to the background (BG), TC, or AR classes.

Specific challenges include the very small dataset size, inherent class imbalance of infrequent extreme events, unavoidable bias due to subjective human labeling, and limited capacity of the light-weight network. We report experiments with different hyper-parameters (loss function, learning rate), architecture (up-sampling), data augmentation, and feature engineering. We find that weighted loss functions aimed at compensating class imbalance provide the most significant improvement on recall of extreme weather events.

\section{Related work}
Initial inspiration for this work came from \citet{ClimateNet} which trained a DeepLabV3+ convolutional neural net on the \textit{ClimateNet} expert-label dataset. This $\sim$50 million parameters model achieved an intersection over union (IoU) score (\ref{eq:iou}) of 0.24 for TCs, and was the first to demonstrate that deep learning models trained on hand-labeled climate data could effectively perform semantic segmentation of extreme weather patterns. However, the DeepLabV3+ architecture is complex, heavy, and thus costly in terms of memory, training time, and associated carbon footprint.

In \textit{Spatio-temporal segmentation and tracking of weather patterns with light-weight Neural Networks}, \citet{NeurIPS2020} attempt to perform the same segmentation task on the \textit{ClimateNet} dataset with the much lighter-weight ($\sim$500,000 parameters) Context Guided neural architecture. They improve on \citet{ClimateNet} with a IoU score of 0.34 and a recall of 0.57 for TCs, our primary class of interest. This model and its associated metrics form our performance baseline.

For a detailed presentation of Context Guided convolutional neural networks, we refer the reader to the original paper that introduced the CGNet architecture, \textit{A light-weight Context Guided Network for semantic segmentation} by \citet{CGNet}. To solve the class imbalance problem, we experimented with various loss functions reviewed in \textit{Survey of loss functions for semantic segmentation} \citep{Jadon_2020}. Lastly, we relied on \textit{Deep Learning for the Earth Sciences} \citep{DL4Earth} for general background on applying deep learning techniques to Earth Sciences.

\section{Dataset \& Features}

\begin{figure}[!h]
\centering
\resizebox{\textwidth}{!}{%
\begin{tabular}{ccc}
%\multicolumn{3}{c}{\textbf{Ground truth labels \& associated sample channels} (TMQ \& U850) } \\  \\
\multicolumn{3}{c}{\textit{North America}} \\
\includegraphics[width=0.33\textwidth]{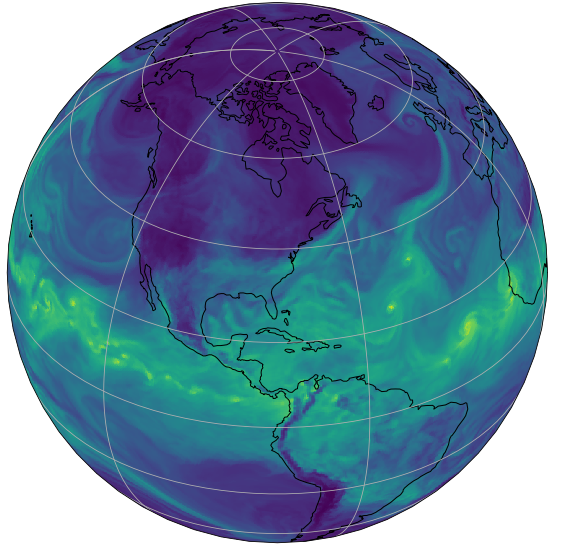} & \includegraphics[width=0.33\textwidth]{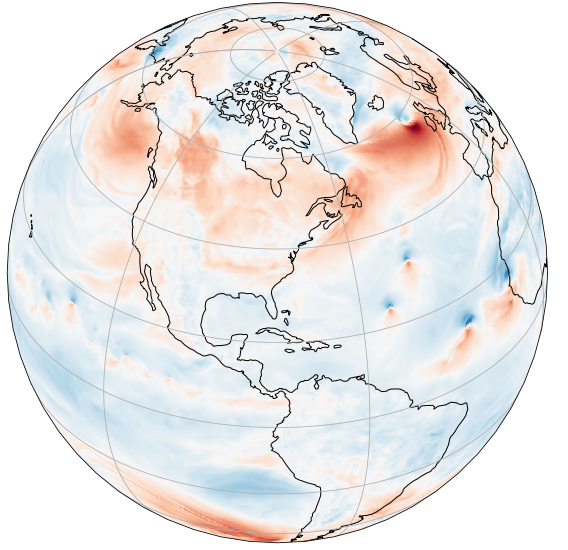} & \includegraphics[width=0.33\textwidth]{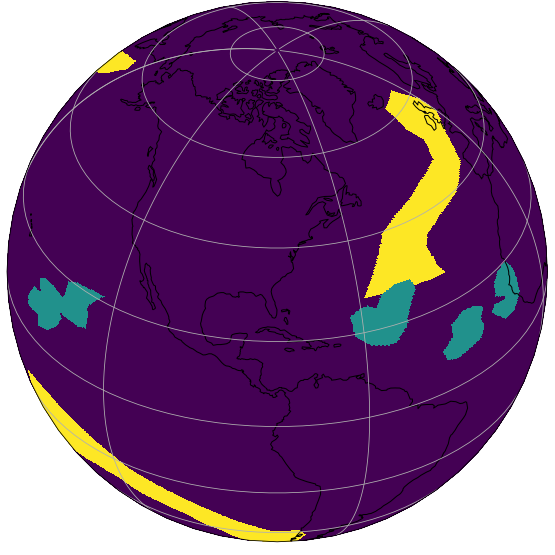} \\
 \\ 
\multicolumn{3}{c}{\textit{Asia-Pacific}} \\
\includegraphics[width=0.33\textwidth]{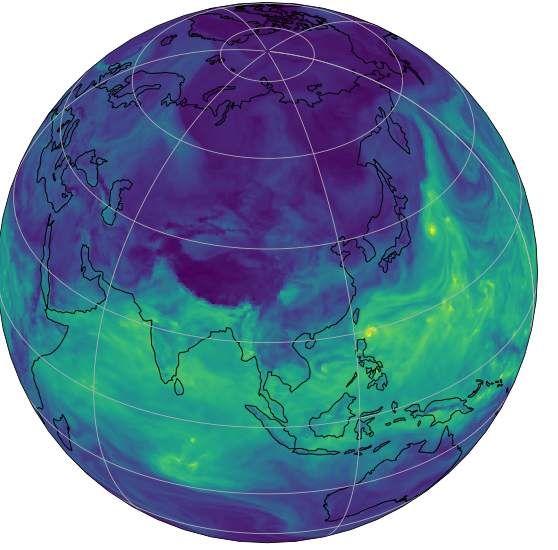} & \includegraphics[width=0.33\textwidth]{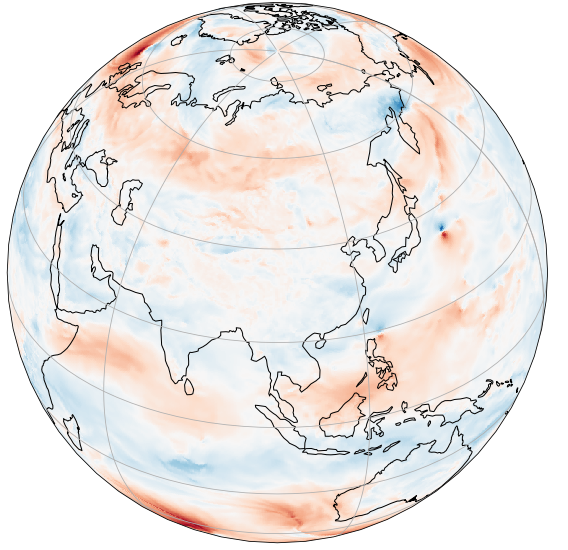} & \includegraphics[width=0.33\textwidth]{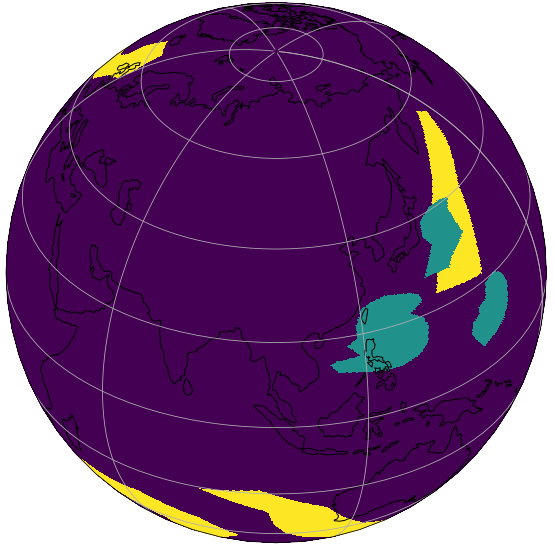}   \\
\end{tabular}}
%\bigskip
\caption{\textbf{Sample \emph{ClimateNet} channels and associated ground truth labels} (TMQ \& U850). \\ AR: yellow; TC: green; BG: purple. Viewed from 35$^{\circ}$N 80$^{\circ}$W (above) and 35$^{\circ}$N  100$^{\circ}$E (below).}
\label{fig:channels-visuals}
\end{figure}

We trained our neural net on \textit{ClimateNet}, an open, community-sourced, human expert-labeled dataset of outputs from Community Atmospheric Model (CAM5.1) climate simulation runs for 459 time steps from 1996 to 2013. Each sample is a netCDF file containing a 1152 $\times$ 768 array for one simulation time step, with each pixel mapping to: one (latitude, longitude) point with 34.8 km/pixel horizontal and 26.1 km/pixel vertical resolution near the Equator; 16 channels for key atmospheric variables, described in table \ref{tab:channels-description} and visualized in figure \ref{fig:channels-visuals}; and one ground truth class label. The dataset is split into a training set of 398 (map, labels) pairs from 1996 to 2010, and a test set of 61 (map, labels) pairs spanning 2011 to 2013. For learning rate scheduling, we created a validation set of 56 (map, labels) pairs spanning 2008 to 2010, which we set aside from the training set to keep the test set consistent with our baseline.

The implementation by \citet{ClimateNetLib} is trained on the following four channels: TMQ, total vertically integrated precipitable water; U850, zonal (east-west) winds at the 850 mbar pressure surface; V850, meridional (north-south) wind at the 850 mbar pressure surface; and PSL, atmospheric pressure at sea level. From the existing 16 channels, we engineered new features, \textit{wind velocity} and \textit{wind vorticity}, to help the model identify TCs since they are characterized by high wind speeds and rotation. Wind velocity is the $L_2$ norm of zonal and meridional components of the wind vector field (equation \ref{eq:VRT}). Wind vorticity is the curl of the wind vector field around the earth radius axis (equation \ref{eq:WS}), a measure of the local rotation \citep{Vorticity}. We pre-computed these engineered features at the 850 mbar pressure level and at the lowest altitude level.

The output of the model is a (1152 $\times$ 768) tensor of softmax probabilities for background, TC, or AR classes. Importantly, labels for the supervised learning of this task are segmentation maps that were hand-drawn by climate scientists as part of a community labeling exercise described in \citet{ClimateNet}. Figure \ref{fig:labeled-data} illustrates how labels were generated as a consensus between experts.

In an effort to reduce over-fitting to the relatively small training set, we explored data augmentation techniques. While transforming the image based on randomized longitude increments seemed promising, we observed that random translations along the longitudes dimension immediately decreased performance. We hypothesize that this may be due to the importance of geography (relative positioning of continents and oceans) for atmospheric circulation and weather patterns. As a consequence, rather than providing additional data for training, data augmentation may act as a detriment to learning by precluding the learning of accurate geographical representations.

\section{Methods} 

\subsection{Baseline Implementation and Performance}

We established our baseline by training the \citet{ClimateNetLib} implementation of the CGNet architecture for 15 epochs over the \textit{ClimateNet} training set, with a Jaccard loss (equation \ref{eq:jaccard-loss}) based on the IoU for the 3 classes (background, AR, and TC).

We report recall as a key performance metric to minimize false negatives, which is especially important for identification of infrequent events. The baseline performance for TCs reaches an IoU score of 0.3396 and a recall of 0.5708 on the test set (see table \ref{tab:results}). A higher performance on the train set (IoU score of 0.38 for TCs) indicates the model may also display some variance and over-fitting.

A fundamental challenge for climate event identification is the inherent imbalance of the data, since, by definition, the extreme events we aim to detect are very rare. We conclude from this analysis that the baseline implementation exhibits high bias, some variance, and relatively low recall.

\subsection{CGNet Architecture}

The light-weight CGNet architecture introduced by \citet{CGNet} follows the principle of ``deep and thin'' and is designed specifically for high-accuracy semantic segmentation while maintaining a small memory footprint. This is advantageous for reducing training time and model complexity. 

\begin{figure}[!h]
  \centering 
  \includegraphics[width=\textwidth]{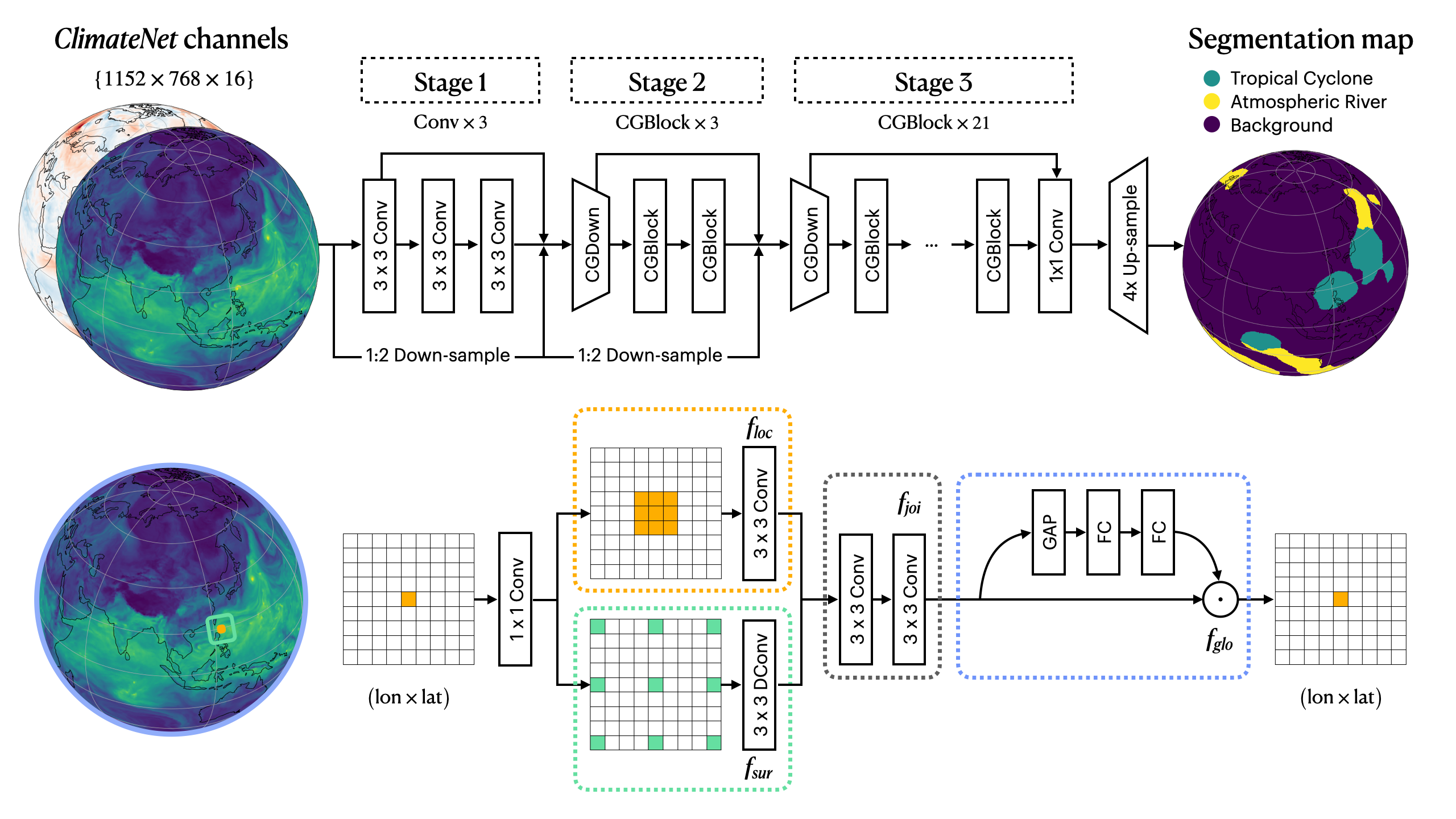}
  \caption{Above: Context Guided convolutional neural network (CGNet). Below: Context Guided block (CG) consisting of local feature extractor $f_{loc}$, surrounding context extractor $f_{sur}$, joint feature extractor $f_{join}$, and global context extractor $f_{glo}$ where $\odot$ represents element-wise multiplication.}
  \label{fig:cgnet-visual}
\end{figure}

\paragraph{Context Guided block.}
The basic unit of CGNet is a Context Guided (CG) block, presented in figure \ref{fig:cgnet-visual}, which concatenates the output of normal and dilated convolutional layers to integrate local and surrounding context respectively. It uses 1x1 convolutions and average pooling to further refine the representation using a global context. The CG block reduces parameter count and memory footprint by employing channel-wise convolutions to lower computational cost across channels. 

\paragraph{Architectural experimentations.}
In order to improve performance, we experimented with additional CNN + BatchNorm + ReLU layers to the model to produce a deeper network with the goal of learning more complex features. We also experimented with doubling the final up-sampling layer to increase resolution of the output predictions. Both of these attempts were unsuccessful at significantly improving performance. 

\paragraph{Learning rate scheduler.}
Experimenting with learning rates greater and lower than the original (0.001) negatively affected IoU and Dice scores. To limit variance, we implemented learning rate (LR) scheduling and early termination for the Adam optimizer. This proved successful in reducing the over-fitting observed in the baseline.

\subsection{Addressing Imbalanced Classes}

The foremost challenge presented by this task is the extreme data imbalance inherent to rare weather events. \citet{ClimateNet} report 94\% of pixels in the \textit{ClimateNet} data belonging to the background class. We find that TCs represent only 0.462\% of pixels of the entire dataset (and ARs only 5.674\%). This means that a naive model assigning \textit{every pixel} to the background class would reach 94\% accuracy despite failing at its task. 

To address this class imbalance, we experimented with modifying the loss landscape to better account for under-represented classes and improve performance on rare events such as TC and AR pixels. To that end, we leaned on the literature review by \citet{Jadon_2020} to  select and implement additional performance metrics and loss functions for training and evaluation.

\subsubsection{Performance metrics} 
To fulfill our problem statement of improved detection of rare weather events in climate data, we explored performance metrics that better represent the model's capacity to learn that task. Specifically, we value detecting extreme events more than identifying their exact boundaries hand-labeled by experts, and aim to penalize missing relevant events more than over-predicting their geographical extent. Specifically, we implemented the following performance metrics:

\begin{itemize}
  \item \textbf{Intersection over union:} our baseline model was trained to optimize for the IoU metric (equation \ref{eq:iou}), as usual for many computer vision problems.
  \item \textbf{Sørensen–Dice similarity} or Dice coefficient (equation \ref{eq:dice}) is a measure of the similarity between class predictions and ground truth that is widely used for image comparison.
  \item \textbf{Recall} or \textbf{Sensitivity:} we devised our training strategy to optimize for recall (equation \ref{eq:recall}) as a proxy for the ability to detect most true positives of the TC class.  
\end{itemize}

\subsubsection{Weighted loss functions} 

To optimize for these metrics, we explored and implemented a broad set of loss functions designed to assign higher weights to rare classes, building on a review by \cite{Jadon_2020}:

\begin{itemize}
\item \textbf{Jaccard loss:} used by our baseline mode. Computes a derivable prediction of segmentation map  IoU from the softmax probabilities output of the classifier (equation \ref{eq:jaccard-loss}).

\item \textbf{Dice loss:} derivable Dice coefficient from the softmax probabilities (equation \ref{eq:dice-loss}).

\item \textbf{Cross-entropy loss:} canonically used in multi-class classification problems, it helps balance under-represented classes. We used the pyTorch implementation of the cross entropy loss (equation \ref{eq:crossentropy-loss}) and weighted cross entropy loss (equation \ref{eq:weighted-crossentropy-loss}).

\item \textbf{Focal Tversky loss:} a tunable loss function which gives higher weight to false positives, false negatives, and hard examples, by introducing hyper-parameters $\beta$ and $\gamma$ (equation \ref{eq:tversky-loss}).

\item \textbf{Weighted Jaccard loss:} to normalize the relative weights of each class in the IoU estimate, we experimented with a custom loss function inspired by the Jaccard loss (equation \ref{eq:weighted-jaccard-loss}).

\end{itemize}

\section{Results \& Discussion}

\begin{figure}[h]
  \includegraphics[width=\textwidth]{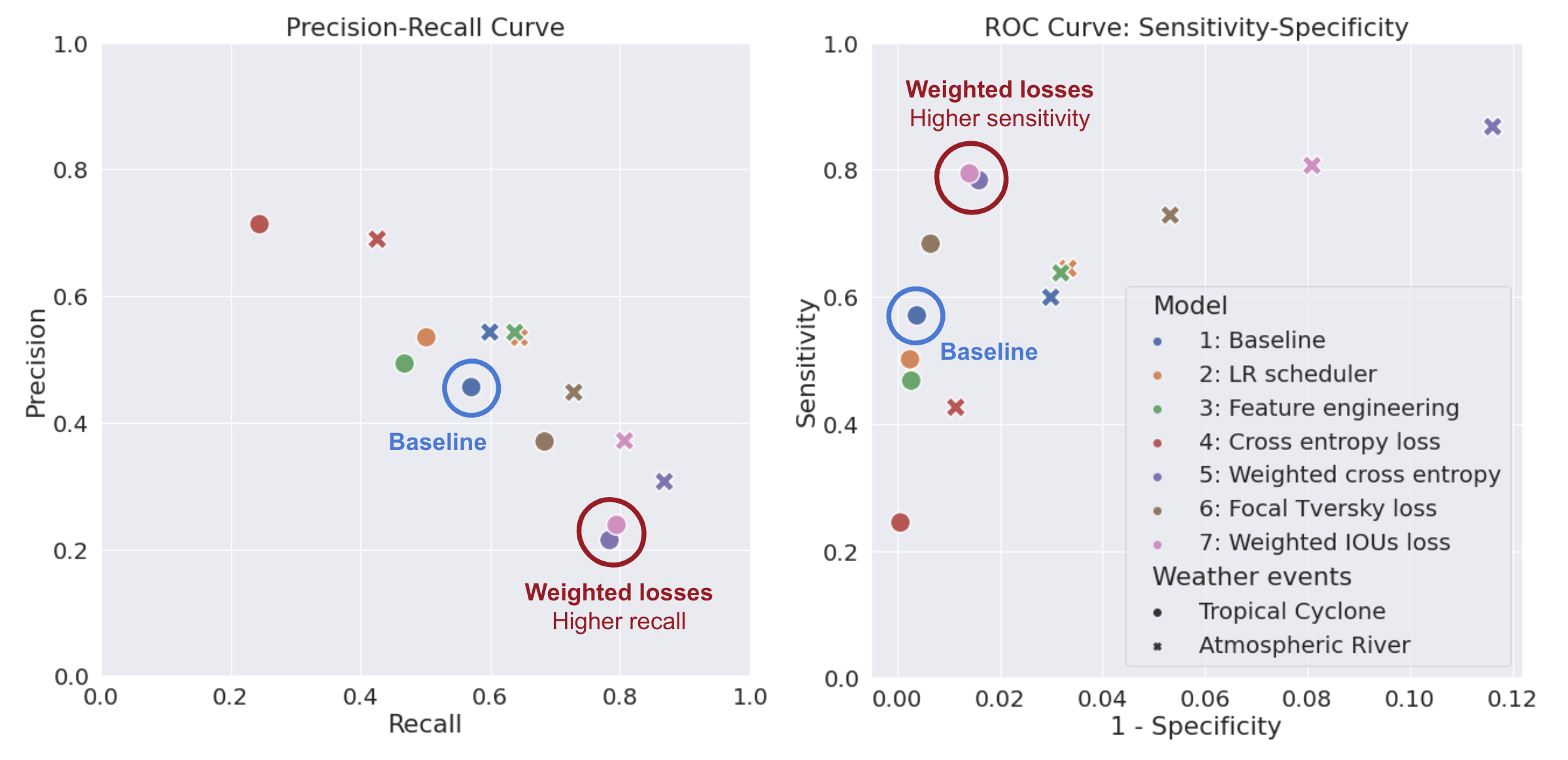}
  \centering 
  \caption{(i) Precision-Recall Curve (left); (ii) ROC Curve (Sensitivity vs 1-Specificity, right).\\ ($\bullet$): tropical cyclone. ($\times$): atmospheric river.}
  \label{fig:results}
\end{figure}

\begin{figure}[!h]
\centering
%\bigskip
\resizebox{\textwidth}{!}{%
\begin{tabular}{ccccc}
%  \\ \hline  \\
%\multicolumn{5}{c}{\textbf{Predicted segmentation maps} } \\
 \\
%\multicolumn{5}{c}{ } &
\textbf{\begin{tabular}[c]{@{}c@{}}1: Baseline\\ (theirs)\end{tabular}}  & \textbf{\begin{tabular}[c]{@{}c@{}}3: Feature \\ engineering (ours)\end{tabular}} & \textbf{\begin{tabular}[c]{@{}c@{}}4. Cross \\ entropy (ours)\end{tabular}} & \textbf{\begin{tabular}[c]{@{}c@{}}5. Weighted cross \\  entropy (ours)\end{tabular}}  & \textbf{\begin{tabular}[c]{@{}c@{}}7. Weighted \\ Jaccard (ours)\end{tabular}} \\ \hline \\

%\begin{tabular}[l]{@{}l@{}}1. Baseline\\ model (theirs) \end{tabular} & \begin{tabular}[l]{@{}l@{}}3. Engineered\\ features (ours)\end{tabular} & \begin{tabular}[l]{@{}l@{}}4. Cross entropy\\loss (ours)\end{tabular} & \begin{tabular}[l]{@{}l@{}}5. Weighted\\CE loss (ours)\end{tabular} & \begin{tabular}[l]{@{}l@{}}7. Weighted\\ Jaccard (ours)\end{tabular} \\ \\ 
\multicolumn{5}{c}{\textit{North America}} \\
\includegraphics[width=0.2\textwidth]{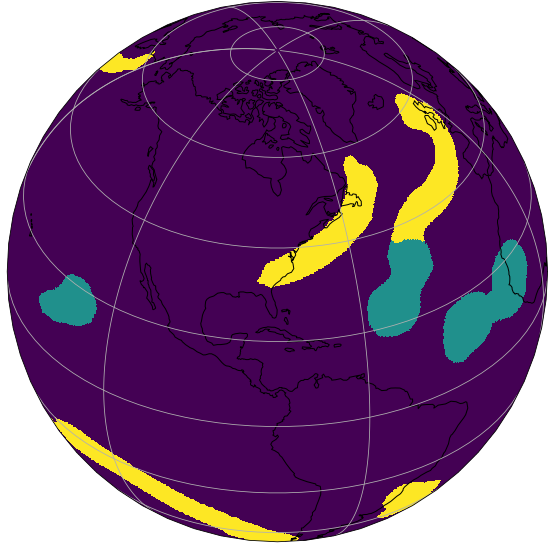}  & \includegraphics[width=0.2\textwidth]{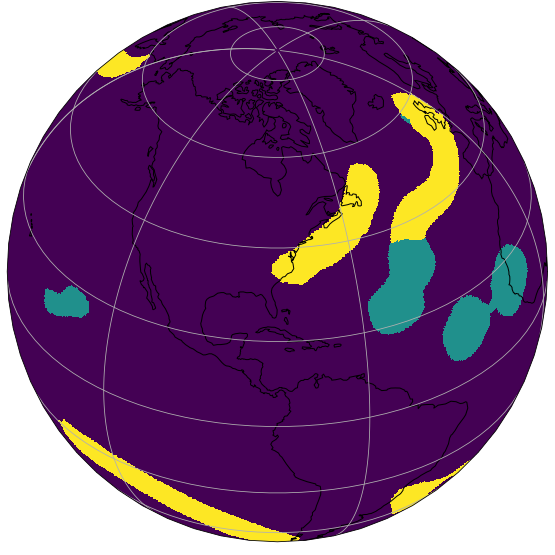} & \includegraphics[width=0.2\textwidth]{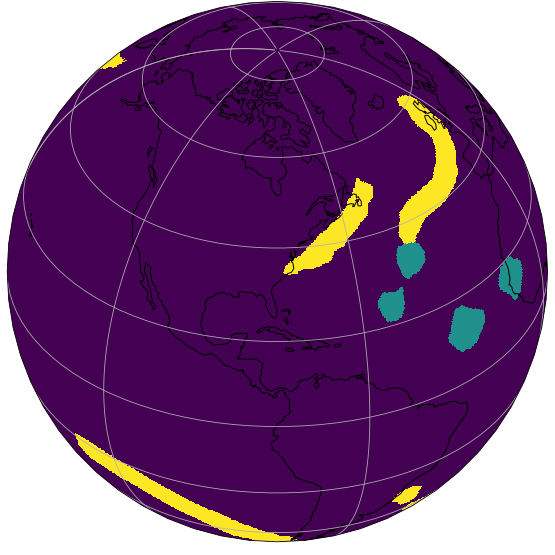}  & \includegraphics[width=0.2\textwidth]{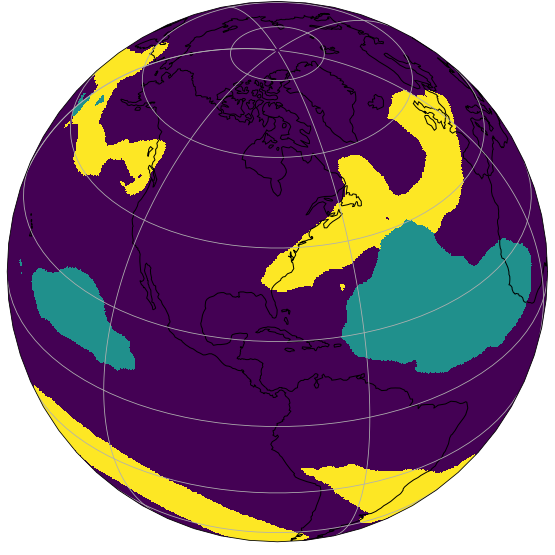} & \includegraphics[width=0.2\textwidth]{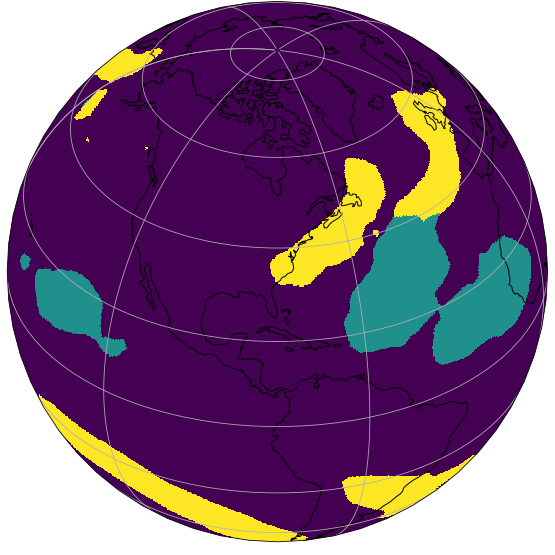}  \\ 
\\
\multicolumn{5}{c}{\textit{Asia-Pacific}} \\
\includegraphics[width=0.2\textwidth]{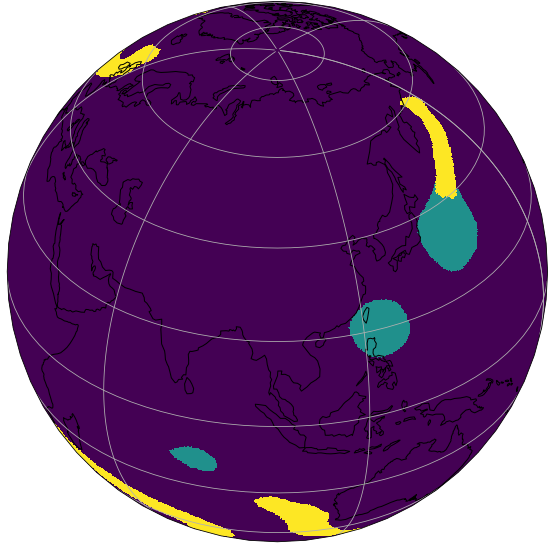} & \includegraphics[width=0.2\textwidth]{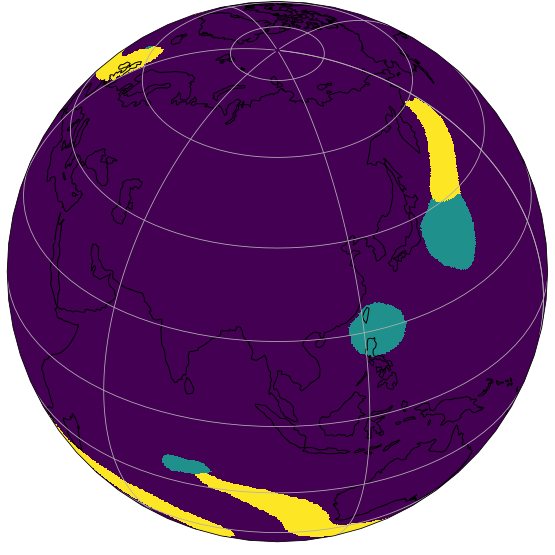} & \includegraphics[width=0.2\textwidth]{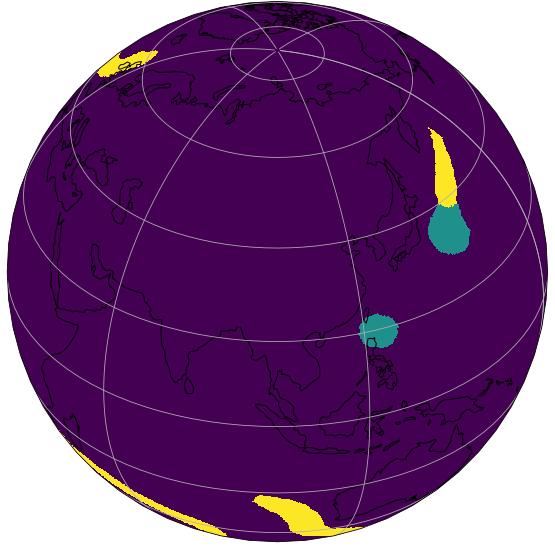} & \includegraphics[width=0.2\textwidth]{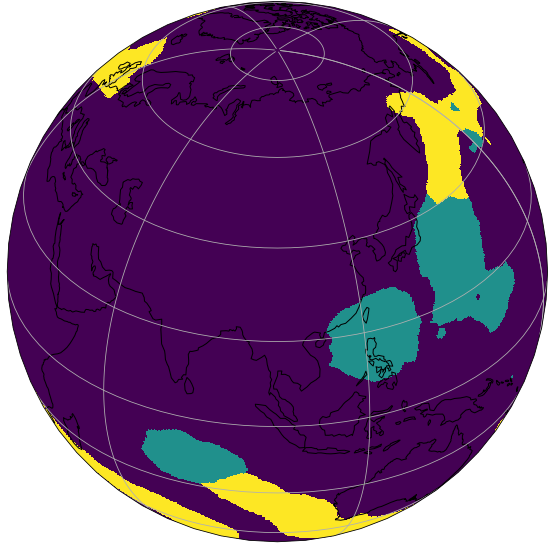} & \includegraphics[width=0.2\textwidth]{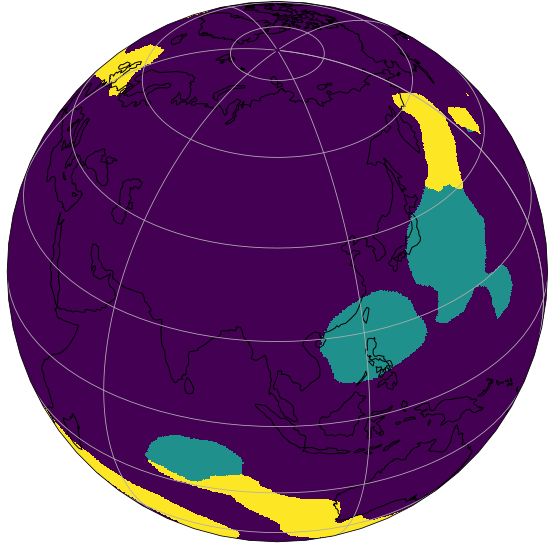}  \\ 
\end{tabular}
}
\caption{\textbf{Predicted segmentation maps} produced by the baseline and four of the models we trained (test set sample). Outputs aim to predict ground truth labels in figure \ref{fig:channels-visuals}.}
\label{fig:metricsvisuals}
\end{figure}

We report summary results for the baseline and six experiments in table \ref{tab:results}, and corresponding precision-recall and ROC curves in figure \ref{fig:results}. Table \ref{tab:results-details} reports detailed performance metrics for our experiments (except data augmentation due to performance drop), and figure \ref{fig:metricsvisuals} compares ground truth labels and baseline results with our predicted segmentation maps on a test set sample.

\begin{table}[!h]
\caption{Summary results for baseline model and six experiments.}
\centering
\resizebox{\textwidth}{!}{%
\begin{tabular}[t]{@{}rlccccccc@{}}
%\toprule
%\multicolumn{9}{c}{ } & 
\multicolumn{2}{l}{\textbf{\begin{tabular}[c]{@{}l@{}} Models\\ \& Metrics\end{tabular}}} & \textbf{\begin{tabular}[c]{@{}c@{}}1: Baseline\\ model\end{tabular}}  & \textbf{\begin{tabular}[c]{@{}c@{}}2: Learning\\ rate decay\end{tabular}} & \textbf{\begin{tabular}[c]{@{}c@{}}3: Feature \\ engineering\end{tabular}} & \textbf{\begin{tabular}[c]{@{}c@{}}4. Cross \\ entropy\end{tabular}} & \textbf{\begin{tabular}[c]{@{}c@{}}5. Weighted \\ cross entropy\end{tabular}} & \textbf{\begin{tabular}[c]{@{}c@{}}6. Focal \\ Tversky\end{tabular}} & \textbf{\begin{tabular}[c]{@{}c@{}}7. Weighted \\ Jaccard\end{tabular}} \\ \hline \\ %\midrule
\textbf{TC} & \textbf{IoU} & 0.3396 & \underline{0.3492} & 0.3161 & 0.2228 & 0.2025 & 0.3160 & 0.2245 \\
\textbf{} & \textbf{Precision} & 0.4560 & 0.5346 & 0.4933 & \underline{0.7134} & 0.2145 & 0.3701 & 0.2384 \\
\textbf{} & \textbf{Recall} & 0.5708 & 0.5016 & 0.4681 & 0.2447 & 0.7836 & 0.6836 & \underline{\textbf{0.7944}} \\
 \textbf{} & \textbf{Specificity} & 0.9962 & 0.9976 & 0.9973 & \underline{0.9995} & 0.9841 & 0.9936 & 0.9860 \\ \\ 
\textbf{AR} & \textbf{IoU} & 0.3983 & 0.4128 & \underline{0.4147} & 0.3575 & 0.2932 & 0.3839 & 0.3411 \\
 \textbf{} & \textbf{Precision} & \underline{0.5429} & 0.5344 & 0.5425 & 0.6896 & 0.3069 & 0.4479 & 0.3714 \\
\textbf{} & \textbf{Recall} & 0.5993 & 0.6448 & 0.6377 & 0.4261 & \underline{0.8680} & 0.7287 & 0.8068 \\
 \textbf{} & \textbf{Specificity} & 0.9701 & 0.9667 & 0.9681 & \underline{0.9886} & 0.8839 & 0.9468 & 0.9191 \\ 
 & & & & & & & & \\
\end{tabular}}
\label{tab:results}
\end{table}

\paragraph{Tropical cyclones recall.}
While we measured IoU, Dice, precision, recall/sensitivity, and specificity scores for TC and AR events, our key results focus on: (i) recall performance to prioritize detection of positives given the severity of a positive event; and (ii) TCs specifically, the most destructive extreme weather events, for which previous models showed limited performance.

\paragraph{Key results.}
After comparing our models on the precision-recall and specificity-sensitivity curves, we found that our weighted Cross Entropy and weighted Jaccard loss models with engineered features and a learning rate scheduler achieve better recall than the baseline (0.7836 and 0.7944 compared to 0.5708, a performance gain of +37.3\% and +39.2\%, respectively). Our experiments with the baseline model with LR scheduler, with baseline loss on engineered data with LR scheduler, and with cross entropy loss on engineered data with LR scheduler performed worse or no better than the baseline (0.2447, 0.4681, and 0.5016, respectively).

\paragraph{Carbon footprint.}
Given the climate focus of this model and our goal of keeping it light-weight, we tracked and evaluated our carbon footprint during our experiments. Based on emissions factors from \citet{co2}, and approximately 40 hours of usage of an NVIDIA A100 GPU VM with 40GB of RAM, we estimate our model training emissions at around 6.24 kg CO$_{2e}$.

\section{Conclusion}

In conclusion, semantic segmentation of extreme weather events in climate data is a challenging problem. The small and imbalanced dataset makes improving on task performance difficult, and CGNet is an intentionally light-weight model with limited capacity. IoU alone is a poor performance metric for identification of rare extreme weather events and should be paired with recall to reflect the priority given to true positive predictions on under-represented classes.

We found success with weighted loss functions, and showed a significant (+39.2\%) improvement in recall for our class of interest. We demonstrated that careful matching of loss functions and optimization algorithms with the task at hand can yield important performance gains, even for light-weight architectures with a much lower resource footprint than current trends in machine learning.

Because advances in light-weight segmentation are so new (the seminal CGNet paper was published in 2021), we have found no other applications of these novel architectures to climate data so far beyond the reported baseline. We hope our results will contribute to improving automated extreme weather events detection, which is of crucial importance to prediction, mitigation, and equitable adaptation to the increasing destructiveness of anthropogenic climate change.

\subsection*{Acknowledgments}
We would like to thank Lukas Kapp-Schwoerer, Andre Graubner, and their co-authors in \citet{NeurIPS2020} for their implementation of CGNet on \textit{ClimateNet} data, the authors of \citet{CGNet} for the original light-weight Context Guided network architecture, and the authors of \citet{ClimateNet} and the climate sciences expert-labeling community for creating and annotating the \textit{ClimateNet} dataset, which made this study possible. We are also grateful to Andrew Ng, Kian Katanforoosh, and Sarthak Consul at Stanford University for their guidance and support. 

\subsection*{Data Availability}
The original \emph{ClimateNet} dataset is available at \url{https://portal.nersc.gov/project/ClimateNet/}. The dataset with engineered features is available at \url{https://huggingface.co/datasets/rlacombe/ClimateNet/}. 

We provide an online repository at \url{https://github.com/hannahg141/ClimateNet} with: (i) our modified implementation of the CGNet model building on \citet{ClimateNetLib}; (ii) notebooks for download, exploration, and visualization of the \textit{ClimateNet} data set, generation of engineered features, and flexible model training on a Google Colab instance; and (iii) a baseline and six experimental models along with their training and evaluation metrics history.

\subsection*{Future Work}
\begin{figure}[!h]
  \centering 
  \includegraphics[width=0.9\textwidth]{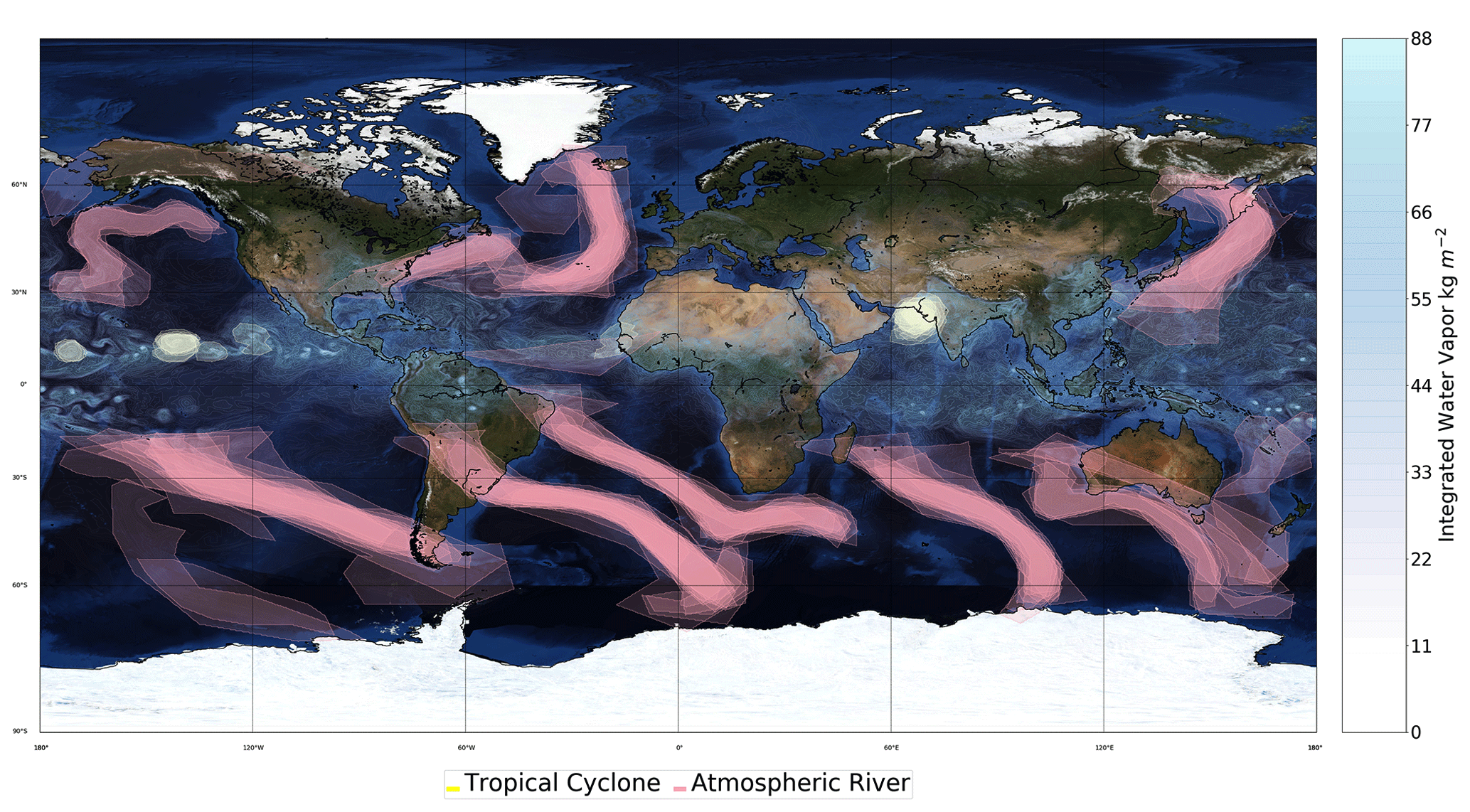}
  \caption{Example image from \citet{ClimateNet} showing 15 different expert labelings (TC labels in white/yellow masks seen near the equator; AR labels in pink masks). The background “blue marble” map included via Matplotlib’s Basemap library is (c) NASA.}
  \label{fig:labeled-data}
\end{figure}

A critical issue with model training on \textit{ClimateNet} is the small and imbalanced nature of the dataset. Also, as is apparent in figure \ref{fig:labeled-data}, individual labels appear to have some degree of subjectivity, and we suspect human-expert consensus labeling leads to unavoidable bias and high Bayes error. Training on historical observational data, expanding expert-labeling efforts, or learning event identification with more objective ground truth labels (e.g. building on previous work on TC centers identification \citep{nguyen2014evaluation}) has the potential to improve performance on this task.

A promising direction for that purpose is the \textit{International Best Track Archive for Climate Stewardship} (IBTrACS) dataset, a historical database of TC positions, wind speeds, and geographical extents maintained by NOAA \citep{IBTrACS}. In conjunction with weather re-analysis data services such as ERA5 \citep{era5}, this set of labels could enable training on a large corpus of observational data. Crucially, the \textit{IBTrACS} data set is global and covers oceanic basins where tropical cyclones with the most destructive impact on LMICs are forming.

This avenue for future work could generalize our models from simulations to observational data, a key step towards early warning and detection systems for equitable adaptation to climate change.

\bibliography{iclr2023_conference}

\begin{thebibliography}{17}
\providecommand{\natexlab}[1]{#1}
\providecommand{\url}[1]{\texttt{#1}}
\expandafter\ifx\csname urlstyle\endcsname\relax
  \providecommand{\doi}[1]{doi: #1}\else
  \providecommand{\doi}{doi: \begingroup \urlstyle{rm}\Url}\fi

\bibitem[{Copernicus Climate Change Service}(2017)]{era5}
{Copernicus Climate Change Service}.
\newblock {ERA5: Fifth generation of ECMWF atmospheric reanalyses of the global
  climate. Copernicus Climate Change Service Climate Data Store (CDS)}.
\newblock 2017.
\newblock URL \url{https://cds.climate.copernicus.eu/cdsapp#!/home}.

\bibitem[IPCC(2022)]{IPCC}
IPCC.
\newblock {Climate Change 2022: Impacts, Adaptation, and Vulnerability.
  Contribution of Working Group II to the Sixth Assessment Report of the
  Intergovernmental Panel on Climate Change}.
\newblock 2022.
\newblock \doi{10.1017/9781009325844}.
\newblock URL \url{https://www.ipcc.ch/report/ar6/wg2/}.

\bibitem[Jadon(2020)]{Jadon_2020}
Shruti Jadon.
\newblock A survey of loss functions for semantic segmentation.
\newblock In \emph{2020 {IEEE} Conference on Computational Intelligence in
  Bioinformatics and Computational Biology ({CIBCB})}. {IEEE}, october 2020.
\newblock \doi{10.1109/cibcb48159.2020.9277638}.
\newblock URL \url{https://doi.org/10.1109%2Fcibcb48159.2020.9277638}.

\bibitem[Kapp-Schwoerer et~al.(2020{\natexlab{a}})Kapp-Schwoerer, Graubner,
  Kim, and Kashinath]{ClimateNetLib}
Lukas Kapp-Schwoerer, Andre Graubner, Sol Kim, and Karthik Kashinath.
\newblock {ClimateNet}, a {Python} library for deep learning-based climate
  science.
\newblock 2020{\natexlab{a}}.
\newblock URL \url{https://github.com/andregraubner/ClimateNet}.

\bibitem[Kapp-Schwoerer et~al.(2020{\natexlab{b}})Kapp-Schwoerer, Graubner,
  Kim, and Kashinath]{NeurIPS2020}
Lukas Kapp-Schwoerer, Andre Graubner, Sol Kim, and Karthik Kashinath.
\newblock Spatio-temporal segmentation and tracking of weather patterns with
  light-weight neural networks.
\newblock \emph{AI for Earth Sciences Workshop at NeurIPS 2020.},
  2020{\natexlab{b}}.
\newblock URL
  \url{https://ai4earthscience.github.io/neurips-2020-workshop/papers/ai4earth_neurips_2020_55.pdf}.

\bibitem[Knapp et~al.(2018)Knapp, Diamond, Kossin, Kruk, and Schreck]{IBTrACS}
K.~R. Knapp, H.~J. Diamond, J.~P. Kossin, M.~C. Kruk, and C.~J. Schreck.
\newblock {International Best Track Archive for Climate Stewardship (IBTrACS)
  Project, Version 4, NOAA National Centers for Environmental Information},
  2018.
\newblock URL
  \url{https://www.ncei.noaa.gov/products/international-best-track-archive}.

\bibitem[Lacoste et~al.(2019)Lacoste, Luccioni, Schmidt, and Dandres]{co2}
Alexandre Lacoste, Alexandra Luccioni, Victor Schmidt, and Thomas Dandres.
\newblock Quantifying the carbon emissions of machine learning, 2019.
\newblock URL \url{https://arxiv.org/abs/1910.09700}.

\bibitem[Mudigonda et~al.(2021)Mudigonda, Ram, Kashinath, Racah, Mahesh, Liu,
  Beckham, Biard, Kurth, Kim, Kahou, Maharaj, Loring, Pal, O'Brien, Kunkel,
  Wehner, and Collins]{DL4Earth}
Mayur Mudigonda, Prabhat Ram, Karthik Kashinath, Evan Racah, Ankur Mahesh,
  Yunjie Liu, Christopher Beckham, Jim Biard, Thorsten Kurth, Sookyung Kim,
  Samira Kahou, Tegan Maharaj, Burlen Loring, Christopher Pal, Travis O'Brien,
  Kenneth~E. Kunkel, Michael~F. Wehner, and William~D. Collins.
\newblock \emph{Deep Learning for the Earth Sciences}.
\newblock John Wiley \& Sons, Ltd, 2021.
\newblock \doi{https://doi.org/10.1002/9781119646181}.
\newblock URL
  \url{https://onlinelibrary.wiley.com/doi/abs/10.1002/9781119646181}.

\bibitem[Nguyen et~al.(2014)Nguyen, Molinari, and Thomas]{nguyen2014evaluation}
Leon~T Nguyen, John Molinari, and Diana Thomas.
\newblock Evaluation of tropical cyclone center identification methods in
  numerical models.
\newblock \emph{Monthly Weather Review}, 142\penalty0 (11):\penalty0
  4326--4339, 2014.

\bibitem[Nguyen et~al.(2013)Nguyen, Robinson, Kaneko, and Komatsu]{econ}
Thanh~Cong Nguyen, Jackie Robinson, Shinji Kaneko, and Satoru Komatsu.
\newblock Estimating the value of economic benefits associated with adaptation
  to climate change in a developing country: A case study of improvements in
  tropical cyclone warning services.
\newblock \emph{Ecological Economics}, 86:\penalty0 117--128, 2013.
\newblock ISSN 0921-8009.
\newblock \doi{https://doi.org/10.1016/j.ecolecon.2012.11.009}.
\newblock URL
  \url{https://www.sciencedirect.com/science/article/pii/S0921800912004508}.
\newblock Sustainable Urbanisation: A resilient future.

\bibitem[{NOAA}(2022)]{NOAA}
{NOAA}.
\newblock National oceanic and atmospheric administration. fast facts:
  Hurricane costs.
\newblock 2022.
\newblock URL
  \url{https://coast.noaa.gov/states/fast-facts/hurricane-costs.html}.

\bibitem[Parks \& Guinto(2022)Parks and Guinto]{TC_LMICs}
Robbie~M. Parks and Renzo~R. Guinto.
\newblock Invited perspective: Uncovering the hidden burden of tropical
  cyclones on public health locally and worldwide.
\newblock \emph{Environmental Health Perspectives}, 130\penalty0 (11):\penalty0
  111306, 2022.
\newblock \doi{10.1289/EHP12241}.
\newblock URL \url{https://ehp.niehs.nih.gov/doi/abs/10.1289/EHP12241}.

\bibitem[Philip et~al.(2020)Philip, Kew, van Oldenborgh, Otto, Vautard, van~der
  Wiel, King, Lott, Arrighi, Singh, and van Aalst]{Attribution}
S.~Philip, S.~Kew, G.~J. van Oldenborgh, F.~Otto, R.~Vautard, K.~van~der Wiel,
  A.~King, F.~Lott, J.~Arrighi, R.~Singh, and M.~van Aalst.
\newblock A protocol for probabilistic extreme event attribution analyses.
\newblock \emph{Advances in Statistical Climatology, Meteorology and
  Oceanography}, 6\penalty0 (2):\penalty0 177--203, 2020.
\newblock \doi{10.5194/ascmo-6-177-2020}.
\newblock URL \url{https://ascmo.copernicus.org/articles/6/177/2020/}.

\bibitem[Prabhat et~al.(2021)Prabhat, Kashinath, Mudigonda, Kim,
  Kapp-Schwoerer, Graubner, Karaismailoglu, von Kleist, Kurth, Greiner, Mahesh,
  Yang, Lewis, Chen, Lou, Chandran, Toms, Chapman, Dagon, Shields, O'Brien,
  Wehner, and Collins]{ClimateNet}
Prabhat, K.~Kashinath, M.~Mudigonda, S.~Kim, L.~Kapp-Schwoerer, A.~Graubner,
  E.~Karaismailoglu, L.~von Kleist, T.~Kurth, A.~Greiner, A.~Mahesh, K.~Yang,
  C.~Lewis, J.~Chen, A.~Lou, S.~Chandran, B.~Toms, W.~Chapman, K.~Dagon, C.~A.
  Shields, T.~O'Brien, M.~Wehner, and W.~Collins.
\newblock {ClimateNet}: an expert-labeled open dataset and deep learning
  architecture for enabling high-precision analyses of extreme weather.
\newblock \emph{Geoscientific Model Development}, 14\penalty0 (1):\penalty0
  107--124, 2021.
\newblock \doi{10.5194/gmd-14-107-2021}.
\newblock URL \url{https://gmd.copernicus.org/articles/14/107/2021/}.

\bibitem[Simpson(2010)]{Vorticity}
Isla Simpson.
\newblock {Circulation and Vorticity (class lecture, Advanced Atmospheric
  Dynamics, University of Toronto)}, 2010.
\newblock URL
  \url{https://www2.cgd.ucar.edu/staff/islas/teaching/3_Circulation_Vorticity_PV.pdf}.

\bibitem[{World Economic Forum}(2022)]{WEF}
{World Economic Forum}.
\newblock \emph{Global Risks Report}.
\newblock 2022.
\newblock URL \url{https://www.weforum.org/reports/global-risks-report-2022/}.

\bibitem[Wu et~al.(2021)Wu, Tang, Zhang, Cao, and Zhang]{CGNet}
Tianyi Wu, Sheng Tang, Rui Zhang, Juan Cao, and Yongdong Zhang.
\newblock {CGNet}: A light-weight context guided network for semantic
  segmentation.
\newblock \emph{IEEE Transactions on Image Processing}, 30:\penalty0
  1169--1179, 2021.
\newblock \doi{10.1109/TIP.2020.3042065}.
\newblock URL \url{https://ieeexplore.ieee.org/document/9292449}.

\end{thebibliography}
\bibliographystyle{iclr2023_conference}

\clearpage

\section*{Appendix}

%\subsection*{ClimateNet Dataset}

\begin{table}[!h]
  \centering
  \caption{\textit{ClimateNet} dataset channels \citep{ClimateNet}.}
  \resizebox{0.9\textwidth}{!}{%
  \begin{tabular}{llr}
     & & \\
    \textbf{CHANNEL} & \textbf{DESCRIPTION} & \textbf{UNITS} \\ 
    \hline \\
    \textbf{TMQ} & Total (vertically integrated) precipitable water & kg/m$^{2}$ \\ 
    \textbf{U850} & Zonal wind at 850 mbar pressure surface & m/s \\ 
    \textbf{V850} & Meridional wind at 850 mbar pressure surface & m/s \\ 
    \textbf{UBOT} & Lowest level zonal wind & m/s \\ 
    \textbf{VBOT} & Lowest model level meridional wind & m/s \\ 
    \textbf{QREFHT} & Reference height humidity & kg/kg \\ 
    \textbf{PS} & Surface pressure & Pa \\ 
    \textbf{PSL} & Sea level pressure & Pa \\ 
    \textbf{T200} & Temperature at 200 mbar pressure surface & K \\ 
    \textbf{T500} & Temperature at 500 mbar pressure surface & K \\ 
    \textbf{PRECT} & Total (convective and large-scale) precipitation rate & m/s \\ 
    \textbf{TS} & Surface temperature (radiative) & K \\ 
    \textbf{TREFHT} & Reference height temperature & K \\ 
    \textbf{Z1000} & Geopotential Z at 1000 mbar pressure surface & m \\ 
    \textbf{Z200} & Geopotential Z at 200 mbar pressure surface & m \\ 
    \textbf{ZBOT} & Lowest model level height & m \\ 
    \textbf{LABELS} & 0: Background, 1: Tropical Cyclone, 2: Atmospheric River & -  \\
        & & \\
    \textbf{Engineered features} & & \\ \hline \\
    \textbf{WS850} & Wind speed (equation \ref{eq:WS}) at 850 mbar pressure surface & m/s \\ 
    \textbf{VRT850} & Relative wind vorticity (eq. \ref{eq:VRT}) at 850 mbar pressure surface & m/s \\ 
    \textbf{WSBOT} & Lowest level wind speed & m/s \\ 
    \textbf{VRTBOT} & Lowest level relative wind vorticity & m/s \\ \\
  \end{tabular}
  }
  \label{tab:channels-description}
\end{table}

\clearpage

%\subsection*{Experiments \& Results}
%  \centering 

\begin{table}[h]
\centering
  \caption{Description of experiments: baseline and six select models we trained.}
    \resizebox{0.9\textwidth}{!}{%
  \begin{tabular}{lllc}
    \\
    \textbf{EXPERIMENT} & \textbf{LOSS FUNCTION}  & \textbf{FEATURES}  & \textbf{LR DECAY}   \\ \hline \\
    \textbf{1. Baseline} & Jaccard loss & Baseline dataset  & No \\ 
    \textbf{2. Learning rate decay} & Jaccard loss & Baseline dataset  & Yes \\ 
    \textbf{3. Feature engineering} & Jaccard loss & Engineered  & Yes \\ 
    \textbf{4. Cross entropy} & Cross-entropy loss & Engineered  & Yes \\ 
    \textbf{5. Weighted cross entropy} & Weighted cross-entropy loss & Engineered  & Yes \\ 
    \textbf{6. Focal Tversky} & Focal Tversky loss & Engineered  & Yes \\ 
    \textbf{7. Weighted Jaccard} & Weighted Jaccard loss & Engineered  & Yes \\ 
  \end{tabular}
  }
  \label{tab:experiments}
\end{table}

\begin{table}[!h]
\caption{Detailed results for baseline model and six experiments.}
\centering
\resizebox{\textwidth}{!}{%
\begin{tabular}[t]{@{}rlccccccc@{}}
%\toprule
%\multicolumn{9}{c}{ } & 
\multicolumn{2}{l}{\textbf{\begin{tabular}[c]{@{}l@{}} Models\\ \& Metrics\end{tabular}}} & \textbf{\begin{tabular}[c]{@{}c@{}}1: Baseline\\ model\end{tabular}}  & \textbf{\begin{tabular}[c]{@{}c@{}}2: Learning\\ rate decay\end{tabular}} & \textbf{\begin{tabular}[c]{@{}c@{}}3: Feature \\ engineering\end{tabular}} & \textbf{\begin{tabular}[c]{@{}c@{}}4. Cross \\ entropy\end{tabular}} & \textbf{\begin{tabular}[c]{@{}c@{}}5. Weighted \\ cross entropy\end{tabular}} & \textbf{\begin{tabular}[c]{@{}c@{}}6. Focal \\ Tversky\end{tabular}} & \textbf{\begin{tabular}[c]{@{}c@{}}7. Weighted \\ Jaccard\end{tabular}} \\ \hline \\ %\midrule
\textbf{TC} & \textbf{IoU} & 0.33955127 & 0.34916790 & 0.31608774 & 0.22278776 & 0.20251324 & 0.31599551 & 0.22453519 \\
\textbf{} & \textbf{Dice} & 0.50696270 & 0.51760482 & 0.48034448 & 0.36439318 & 0.33681665 & 0.48023798 & 0.36672721 \\
\textbf{} & \textbf{Precision} & 0.45598923 & 0.53463995 & 0.49327914 & 0.71342764 & 0.21450748 & 0.37011321 & 0.23838463 \\
\textbf{} & \textbf{Recall} & 0.57076677 & 0.50162175 & 0.46807083 & 0.24468465 & 0.78363352 & 0.68365510 & 0.79444291 \\
 \textbf{} & \textbf{Specificity} & 0.99623709 & 0.99758723 & 0.99734295 & 0.99945687 & 0.98414286 & 0.99357050 & 0.98597405 \\ \\ 
\textbf{AR} & \textbf{IoU} & 0.39832633 & 0.41285328 & 0.41468760 & 0.35750965 & 0.29317686 & 0.38387118 & 0.34108058 \\
 \textbf{} & \textbf{Dice} & 0.56971870 & 0.58442485 & 0.58626032 & 0.52671397 & 0.45342113 & 0.55477878 & 0.50866530 \\
 \textbf{} & \textbf{Precision} & 0.54289729 & 0.53435760 & 0.54252858 & 0.68965182 & 0.30685734 & 0.44789329 & 0.37141691 \\
\textbf{} & \textbf{Recall} & 0.59932803 & 0.64484427 & 0.63766037 & 0.42605401 & 0.86800497 & 0.72866865 & 0.80679887 \\
 \textbf{} & \textbf{Specificity} & 0.97010985 & 0.96671544 & 0.96815083 & 0.98864332 & 0.88386163 & 0.94679579 & 0.91912138 \\ 
 & & & & & & & &  \\
\end{tabular}}
\label{tab:results-details}
\end{table}

\clearpage

%\subsection*{Segmentation Maps}

\clearpage

\subsection*{Equations}

\subsubsection*{Metrics}
Performance metrics for a single sample.\\
Formalism: TP = True Positives, FP = False Positives, TN = True Negatives, FN = False Negatives.

    \begin{equation} \label{eq:iou}
        \textsc{Intersection over Union} = \frac{\textsc{TP}}{\textsc{TP}+\textsc{FP}+\textsc{FN}}
     \end{equation}

    \begin{equation} \label{eq:dice}
        \textsc{Sørensen–Dice similarity} = \frac{2 \textsc{TP}}{2 \textsc{TP}+\textsc{FP}+\textsc{FN}}
    \end{equation}

    \begin{equation} \label{eq:recall}
        \textsc{Recall/Sensitivity} = \frac{\textsc{TP}}{\textsc{TP}+\textsc{FN}}
    \end{equation}

\subsubsection*{Loss functions}
Loss functions for a single sample. Formalism: $y = y_{ij}$ is a one-hot encoded ground truth tensor for the three classes at longitude and latitude $(i, j)$, and $\hat{y} = \hat{y}_{ij}$ is the 3-classes probabilities tensor computed as the softmax of the logits predicted by the network. Parameters are $w_C$, the tensor of weights used to balance under-represented classes, and $\beta$ and $\gamma$, scalars which allow for the tuning of relative weights of false positives and false negatives and of hard examples in the focal Tversky loss. All operations here are element-wise.

    \begin{equation} \label{eq:jaccard-loss}
        \textsc{Jaccard loss}(y, \hat{y}) = 1 - \frac{\hat{y} y}{(\hat{y} + y) - \hat{y} y}         
    \end{equation}

    \begin{equation} \label{eq:dice-loss}
        \textsc{Dice loss}(y, \hat{y}) = 1 - \frac{{2y \hat{y} + 1}}{{y + \hat{y} + 1}}
    \end{equation}
 
    \begin{equation} \label{eq:crossentropy-loss}
        \textsc{Cross Entropy loss}(y, \hat{y}) = - y \log (\hat{y})
    \end{equation}

    \begin{equation} \label{eq:weighted-crossentropy-loss}
        \textsc{Weighted Cross Entropy loss}(y, \hat{y}) = - w_C y \log (\hat{y})
    \end{equation}
      
    \begin{equation} \label{eq:tversky-loss}
        \textsc{Focal Tversky loss}(y,\hat y) = \left(1 - \frac{{y \hat y}}{{\beta (1 - y)\hat y + (1 - \beta )y(1 - \hat y)}}\right)^{\gamma}
    \end{equation}

     \begin{equation} \label{eq:weighted-jaccard-loss}
        \textsc{Weighted Jaccard loss}(y, \hat{y}) = 1 - w_C \frac{\hat{y} y}{(\hat{y} + y) - \hat{y} y}        
    \end{equation}

  \subsubsection*{Wind Velocity}
  Wind speed is the $L_2$ norm of the zonal and meridional components of the wind vector field: 
  
    \begin{equation} \label{eq:WS}
       w_s = \sqrt{u^2 + v^2} 
    \end{equation}

  \subsubsection*{Relative Wind Vorticity}
  Wind vorticity is the rotation of the wind vector field, where $\lambda$ = longitude and $\phi$ = latitude:

\begin{equation} \label{eq:VRT}
  \zeta = \frac{\partial u}{\partial \lambda} - \frac{1}{\cos \phi} \frac{\partial v \cos \phi}{\partial \phi}
\end{equation}

\end{document}